\documentclass[11pt,a4paper]{article}
\usepackage[hyperref]{naaclhlt2019}
\usepackage{times}
\usepackage{latexsym}

\usepackage{url}

\usepackage{xspace}
\usepackage{booktabs}
\usepackage[mathscr]{euscript}
\usepackage{amsmath}
\usepackage{amssymb}
\usepackage{multirow}
\usepackage{graphicx}
\usepackage{enumitem}
\usepackage{subcaption}
\usepackage{relsize}
\usepackage{bbm}
\usepackage{xargs}
\PassOptionsToPackage{pdftex}{xcolor}
\PassOptionsToPackage{dvipsnames}{xcolor}
\usepackage{arydshln}

\newcommand{\acldata}{ACL-ARC\xspace}
\newcommand{\ourdata}{SciCite\xspace}
\newcommand{\calD}{\mathcal{D}}
\newcommand{\calL}{\mathcal{L}}
\newcommand{\V}[1][\mathbf]{#1}
\newcommand{\I}[1][\textit]{#1}
\newcommand{\R}[1]{\mathcal{R}^{#1}}
\newcommand{\Rtwo}[2]{\mathcal{R}^{(#1, #2)}}

\newcommand{\background}{\textsc{Background}\xspace}
\newcommand{\use}{\textsc{Use}\xspace}
\newcommand{\motivation}{\textsc{Motivation}\xspace}

\newcommand{\compare}{\textsc{Compare}\xspace}
\newcommand{\future}{\textsc{FutureWork}\xspace}
\newcommand{\method}{\textsc{Method}\xspace}
\newcommand{\result}{\textsc{ResultComparison}\xspace}

\usepackage[colorinlistoftodos,prependcaption,textsize=scriptsize]{todonotes}
\newcommandx{\unsure}[2][1=]{\todo[linecolor=red,backgroundcolor=red!25,bordercolor=red,#1]{#2}}
\newcommandx{\change}[2][1=]{\todo[linecolor=blue,backgroundcolor=blue!25,bordercolor=blue,#1]{#2}}
\newcommandx{\info}[2][1=]{\todo[linecolor=OliveGreen,backgroundcolor=OliveGreen!25,bordercolor=OliveGreen,#1]{#2}}
\newcommandx{\improvement}[2][1=]{\todo[linecolor=yellow,backgroundcolor=yellow!35,bordercolor=yellow,#1]{#2}}
\newcommandx{\thiswillnotshow}[2][1=]{\todo[disable,#1]{#2}}

\DeclareMathOperator{\softmax}{softmax}

\aclfinalcopy %
\def\aclpaperid{2017} %

\title{Structural Scaffolds for Citation \\Intent  Classification in Scientific Publications}

\author{Arman Cohan \quad\quad Waleed Ammar \quad\quad Madeleine van Zuylen \quad\quad Field Cady \vspace{6pt} \\
  Allen Institute for Artificial Intelligence \\ \vspace{4pt}
  \small{\tt{\{armanc,waleeda,madeleinev,fieldc\}@allenai.org}}
  }

\date{}

\begin{document}
\maketitle
\begin{abstract}
Identifying the intent of a citation in scientific papers (e.g., \I{background information, use of methods, comparing results}) is critical for machine reading of individual publications and automated analysis of the scientific literature.
We propose structural scaffolds, a multitask model to incorporate structural information of scientific papers into citations for effective classification of citation intents.
Our model achieves a new state-of-the-art on an existing ACL anthology dataset (ACL-ARC) with a 13.3\% absolute increase in F1 score, without relying on external linguistic resources or hand-engineered features as done in existing methods.
In addition, we introduce a new dataset of citation intents (SciCite) which is more than five times larger and covers multiple scientific domains compared with existing datasets. Our code and data are available at: \url{https://github.com/allenai/scicite}.

\end{abstract}

\section{Introduction}

\begin{figure}[t]
\centering
\includegraphics[width=0.7\linewidth]{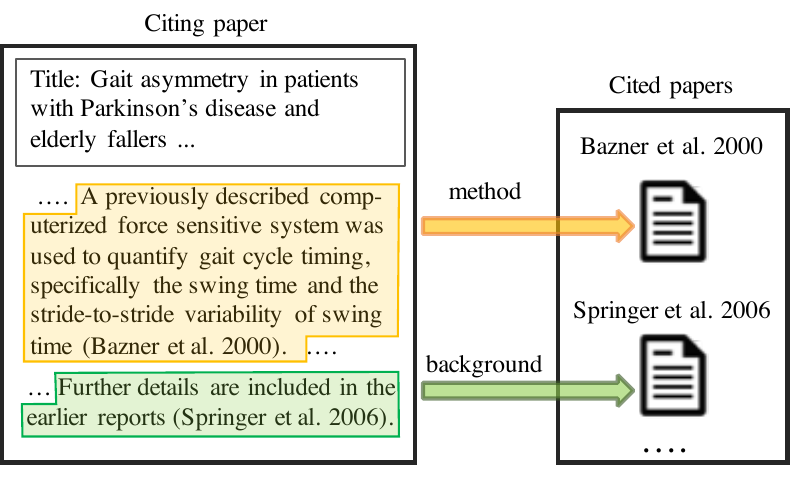}
\caption{
\small{Example of citations with different intents (\background and \method).}
}
\label{fig:problem-definition}
\end{figure}

Citations play a unique role in scientific discourse and are crucial for understanding and analyzing scientific work \cite{luukkonen1992scientists,leydesdorff1998theories}. They are also typically used as the main measure for assessing impact of scientific publications, venues, and researchers \cite{Li2008}.
The nature of citations can be different. Some citations indicate direct use of a method while some others merely serve as acknowledging a prior work.
Therefore, identifying the intent of citations (Figure~\ref{fig:problem-definition}) is critical in improving automated analysis of academic literature and scientific impact measurement \cite{leydesdorff1998theories,small2018}. Other applications of citation intent classification are enhanced research experience \cite{moravcsik1975some}, information retrieval \cite{ritchie2009citation}, summarization \cite{cohan2015summarization}, and studying evolution of scientific fields \cite{jurgens2018}.

In this work, we approach the problem of citation intent classification by modeling the language expressed in the citation context.
A citation context includes text spans in a citing paper describing a referenced work and has been shown to be the primary signal in intent classification \cite{teufel2006function,abu2013purpose,jurgens2018}. Existing models for this problem are feature-based, modeling the citation context with respect to a set of predefined hand-engineered features (such as linguistic patterns or cue phrases) and ignoring other signals that could improve prediction.

In this paper we argue that better representations can be obtained directly from data, sidestepping problems associated with external features. To this end, we propose a neural multitask learning framework to incorporate knowledge into citations from the structure of scientific papers. In particular, we propose two auxiliary tasks as \I{structural scaffolds} to improve citation intent prediction:\footnote{We borrow the scaffold terminology from Swayamdipta et al. (2018) in the context of multitask learning.} (1) predicting the section title in which the citation occurs and (2) predicting whether a sentence needs a citation.
Unlike the primary task of citation intent prediction, it is easy to collect large amounts of training data for scaffold tasks since the labels naturally occur in the process of writing a paper and thus, there is no need for manual annotation. On two datasets, we show that the proposed neural scaffold model outperforms existing methods by large margins.

Our contributions are: \textit{(i)} we propose a neural scaffold framework for citation intent classification to incorporate into citations knowledge from structure of scientific papers; \textit{(ii)} we achieve a new state-of-the-art of 67.9\% F1 on the ACL-ARC citations benchmark, an absolute 13.3\% increase over the previous state-of-the-art \cite{jurgens2018}; and \textit{(iii)} we introduce \ourdata, a new dataset of citation intents which is at least five times as large as existing datasets and covers a variety of scientific domains.

\section{Model}
\label{sec:model}

\begin{figure}[t]
\centering
\includegraphics[width=\linewidth]{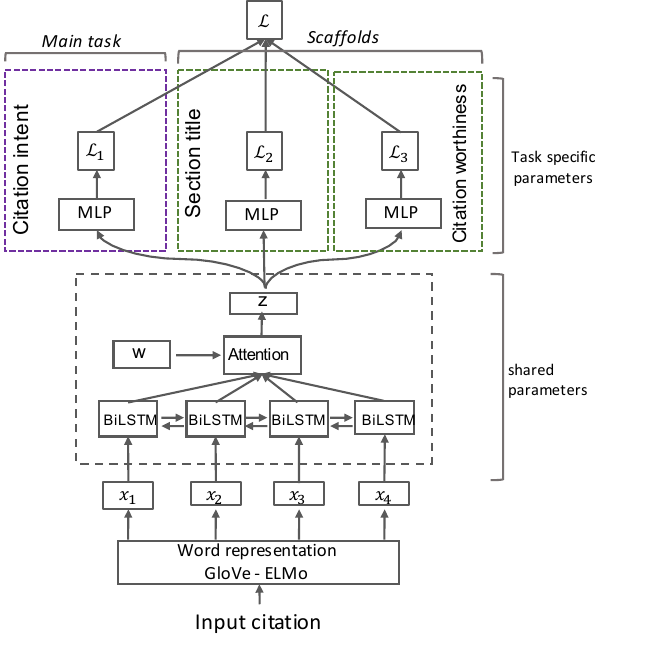}
\caption{\small{Our proposed scaffold model for identifying citation intents. The main task is predicting the citation intent (top left) and two scaffolds are predicting the section title and predicting if a sentence needs a citation (citation worthiness).}
}
\label{fig:model}
\end{figure}

We propose a neural multitask learning framework for classification of citation intents. In particular, we introduce and use two structural scaffolds, auxiliary tasks related to the structure of scientific papers.
The auxiliary tasks may not be of interest by themselves but are used to inform the main task. Our model uses a large auxiliary dataset to incorporate this structural information available in scientific documents into the citation intents. The overview of our model is illustrated in Figure~\ref{fig:model}.

Let $C$ denote the citation and $\V{x}$ denote the citation context relevant to $C$.
We encode the tokens in the citation context of size $n$ as $\V{x}=\{\V{x}_1, ..., \V{x}_n\}$, where $\V{x}_i\in\R{d_1}$ is a word vector of size $d_1$ which concatenates non-contextualized word representations \cite[GloVe,][]{pennington2014glove} and contextualized embeddings \cite[ELMo,][]{Peters2018DeepCW}, i.e.:
$$\V{x}_i = \big[\V{x}_i^{\text{GloVe}};\V{x}_i^{\text{ELMo}}\big]$$
We then use a bidirectional long short-term memory \cite{Hochreiter1997LongSM} (BiLSTM) network with hidden size of $d_2$ to obtain a contextual representation of each token vector with respect to the entire sequence:\footnote{In our experiments BiGRUs resulted in similar performance.}
$$ \V{h}_i = \big[\overrightarrow{\mathrm{LSTM}}(\V{x}, i);\overleftarrow{\mathrm{LSTM}}(\V{x}, i)\big],$$
where $ \V{h} \in \Rtwo{n}{2d_2} $ and $\overrightarrow{\mathrm{LSTM}}(\V{x},i)$ processes $\V{x}$ from left to write and returns the LSTM hidden state at position $i$ (and vice versa for the backward direction $\overleftarrow{\mathrm{LSTM}}$).
We then use an attention mechanism to get a single vector representing the whole input sequence:
$$ \V{z} = \sum_{i=1}^n\alpha_i \V{h}_i, \quad \alpha_i = \softmax(\V{w}^\top\V{h}_i),$$
where $\V{w}$ is a parameter served as the query vector for dot-product attention.\footnote{We also experimented BiLSTMs without attention; we found that BiLSTMs/BiGRUs along with attention provided best results. Other types of attention such as additive attention result in similar performance.} So far we have obtained the citation representation as a vector~$\V{z}$. Next, we describe our two proposed structural scaffolds for citation intent prediction.

\subsection{Structural scaffolds}
\label{subsec:mtl}

In scientific writing there is a connection between the structure of scientific papers and the intent of citations.
To leverage this connection for more effective classification of citation intents, we propose a multitask framework with two structural scaffolds (auxiliary tasks) related to the structure of scientific documents. A key point for our proposed scaffolds is that they do not need any additional manual annotation as labels for these tasks occur naturally in scientific writing. The structural scaffolds in our model are the following:

\paragraph{Citation worthiness.}
The first scaffold task that we consider is ``citation worthiness'' of a sentence, indicating whether a sentence needs a citation. The language expressed in citation sentences is likely distinctive from regular sentences in scientific writing, and such information could also be useful for better language modeling of the citation contexts.
To this end, using citation markers such as ``[12]'' or ``Lee et al (2010)'', we identify sentences in a paper that include citations and the negative samples are sentences without citation markers.
The goal of the model for this task is to predict whether a particular sentence needs a citation.\footnote{We note that this task may also be useful for helping authors improve their paper drafts. However, this is not the focus of this work.}

\paragraph{Section title.}

The second scaffold task relates to predicting the section title in which a citation appears. Scientific documents follow a standard structure where the authors typically first introduce the problem, describe methodology, share results, discuss findings and conclude the paper. The intent of a citation could be relevant to the section of the paper in which the citation appears. For example, method-related citations are more likely to appear in the methods section. Therefore, we use the section title prediction as a scaffold for predicting citation intents.
Note that this scaffold task is different than simply adding section title as an additional feature in the input. We are using the section titles from a larger set of data than training data for the main task as a proxy to learn linguistic patterns that are helpful for citation intents.
In particular, we leverage a large number of scientific papers for which the section information is known for each citation to automatically generate large amounts of training data for this scaffold task.\footnote{We also experimented with adding section titles as additional feature to the input, however, it did not result in any improvements.}

\paragraph{Multitask formulation.}
Multitask learning as defined by \citet{Caruana1997MultitaskL} is an approach to inductive transfer learning that improves generalization by using the domain information contained in the training signals of related tasks as an inductive bias. It requires the model to have at least some sharable parameters between the tasks. In a general setting in our model, we have a main task $Task^{(1)}$ and $n-1$ auxiliary tasks $Task^{(i)}$. As shown in Figure~\ref{fig:model}, each scaffold task will have its task-specific parameters for effective classification and the parameters for the lower layers of the network are shared across tasks. We use a Multi Layer Perceptron (MLP) for each task and then a softmax layer to obtain prediction probabilites. In particular, given the vector $\V{z}$ we pass it to $n$ MLPs and obtain $n$ output vectors $\V{y}^{(i)}$:
$$ \V{y}^{(i)} = \softmax(\mathrm{MLP}^{(i)}(\V{z})) $$

We are only interested in the output $\V{y}^{(1)}$ and the rest of outputs $(\V{y}^{(2)}, ..., \V{y}^{(n)})$ are regarding the scaffold tasks and only used in training to inform the model of knowledge in the structure of the scientific documents. For each task, we output the class with the highest probability in $\V{y}$. An alternative inference method is to sample from the output distribution.

\setlength{\dashlinedash}{0.5pt}
\setlength{\dashlinegap}{1.0pt}
\setlength{\arrayrulewidth}{0.1pt}

\begin{table*}[t]
\scriptsize
\centering
\renewcommand{\arraystretch}{1.2}
\setlength{\tabcolsep}{1pt}
\begin{tabular}{@{}lll@{}}
\toprule
Intent cateogry & Definition & Example \\ \midrule
\begin{tabular}[c]{@{}l@{}}Background\\ information \end{tabular} & \begin{tabular}[c]{@{}l@{}}The citation states, mentions, or points to the background \\ information giving more context about a problem, concept,\\ approach, topic, or importance of the problem in the field. \end{tabular} & \begin{tabular}[c]{@{}l@{}}Recent evidence suggests that co-occurring alexithymia may explain deficits {[}12{]}.\\ Locally high-temperature melting regions can act as permanent termination sites {[}6-9{]}.\\ One line of work is focused on changing the objective function (Mao et al., 2016).\end{tabular} \\ \hdashline
Method & \begin{tabular}[c]{@{}l@{}}Making use of a method, tool, approach or dataset\end{tabular} & \begin{tabular}[c]{@{}l@{}}Fold differences were calculated by a mathematical model described in {[}4{]}.\\ We use Orthogonal Initialization (Saxe et al., 2014)\end{tabular} \\ \hdashline
\begin{tabular}[c]{@{}l@{}}Result \\ comparison \end{tabular} & \begin{tabular}[c]{@{}l@{}}Comparison of the paper's results/findings with the \\ results/findings of other work\end{tabular} & \begin{tabular}[c]{@{}l@{}}Weighted measurements were superior to T2-weighted contrast imaging which was in \\ accordance with former studies [25-27]\\ Similar results to our study were reported in the study of Lee et al (2010).\end{tabular} \\ \bottomrule
\end{tabular}
\caption{The definition and examples of citation intent categories in our \ourdata.}
\label{tab:data-categories}
\end{table*}

\begin{table}[t]
\scriptsize
\centering
\setlength{\tabcolsep}{2pt}
\renewcommand{\arraystretch}{1.3}
\begin{tabular}{@{}lllrr@{}}
\toprule
Dataset & \begin{tabular}[t]{@{}l@{}}Categories\\ (distribution)\end{tabular} & Source & \#papers & \#instances \\ \midrule
ACL-ARC & \begin{tabular}[t]{@{}l@{}}Background (0.51)\\ Extends (0.04)\\ Uses (0.19)\\ Motivation (0.05)\\ Compare/Contrast (0.18)\\ Future work (0.04)\end{tabular} & \begin{tabular}[t]{@{}l@{}}Computational\\ Linguistics\end{tabular} & 186 & 1,941 \\ \hdashline
\ourdata & \begin{tabular}[t]{@{}l@{}}Background (0.58)\\ Method (0.29)\\ Result comparison (0.13)\end{tabular} & \begin{tabular}[t]{@{}l@{}}Computer \\ Science \&\\ Medicine\end{tabular} & 6,627 & 11,020 \\ \bottomrule
\end{tabular}
\caption{Characteristics of \ourdata compared with \acldata dataset by \citet{jurgens2018}
}
\label{tab:data}
\end{table}

\subsection{Training}
\label{subsec:training}

Let $\mathcal{D}_1$ be the labeled dataset for the main task $Task^{(1)}$, and $\mathcal{D}_i$ denote the labeled datasets corresponding to the scaffold task $Task^{(i)}$ where $i\in\{2,...,n\}$. Similarly, let $\mathcal{L}_1$ and  $\mathcal{L}_i$ be the main loss and the loss of the auxiliary task $i$, respectively. The final loss of the model is:
\begin{equation}
\small
\label{eq:loss}
\calL=\sum_{(\V{x},\V{y})\in \calD_1} \calL_1(\V{x},\V{y}) + \sum_{i=2}^n \lambda_i \sum_{(\V{x},\V{y})\in \calD_i} \calL_i(\V{x},\V{y}),
\end{equation}
where $\lambda_i$ is a hyper-parameter specifying the sensitivity of the parameters of the model to each specific task. Here we have two scaffold tasks and hence $n{=}3$. $\lambda_i$ could be tuned based on performance on validation set (see \S\ref{sec:experiments} for details).

We train this model jointly across tasks and in an end-to-end fashion.
In each training epoch, we construct mini-batches with the same number of instances from each of the $n$ tasks. We compute the total loss for each mini-batch as described in Equation~\ref{eq:loss}, where $\calL_i{=}0$ for all instances of other tasks $j{\neq}i$.
We compute the gradient of the loss for each mini-batch and tune model parameters using the AdaDelta optimizer \cite{zeiler2012adadelta} with gradient clipping threshold of 5.0.
We stop training the model when the development macro F1 score does not improve for five consecutive epochs.

\section{Data}
\label{sec:data}

We compare our results on two datasets from different scientific domains. While there has been a long history of studying citation intents, there are only a few existing publicly available datasets on the task of citation intent classification. We use the most recent and comprehensive (\acldata citations dataset) by \citet{jurgens2018} as a benchmark dataset to compare the performance of our model to previous work. In addition, to address the limited scope and size of this dataset, we introduce \ourdata, a new dataset of citation intents that addresses multiple scientific domains and is more than five times larger than \acldata.
Below is a description of both datasets.

\subsection{ACL-ARC citations dataset}
\acldata is a dataset of citation intents released by \citet{jurgens2018}. The dataset is based on a sample of papers from the ACL Anthology Reference Corpus \cite{Bird2008TheAA} and includes 1,941 citation instances from 186 papers and is annotated by domain experts in the NLP field. The data was split into three standard stratified sets of train, validation, and test with 85\% of data used for training and remaining 15\% divided equally for validation and test. Each citation unit includes information about the immediate citation context, surrounding context, as well as information about the citing and cited paper. The data includes six intent categories outlined in Table~\ref{tab:data}.

\subsection{\ourdata dataset}
Most existing datasets contain citation categories that are too fine-grained. Some of these intent categories are very rare or not useful in meta analysis of scientific publications. Since some of these fine-grained categories only cover a minimal percentage of all citations, it is difficult to use them to gain insights or draw conclusions on impacts of papers. Furthermore, these datasets are usually domain-specific and are relatively small (less than 2,000 annotated citations).

To address these limitations, we introduce \ourdata, a new dataset of citation intents that is significantly larger, more coarse-grained and general-domain compared with existing datasets. Through examination of citation intents, we found out many of the categories defined in previous work such as motivation, extension or future work, can be considered as background information providing more context for the current research topic. More interesting intent categories are a direct use of a method or comparison of results. Therefore, our dataset provides a concise annotation scheme that is useful for navigating research topics and machine reading of scientific papers. We consider three intent categories outlined in Table~\ref{tab:data-categories}: \background, \method and \result. Below we describe data collection and annotation details.

\subsubsection{Data collection and annotation}
Citation intent of sentence extractions was labeled through the crowdsourcing platform Figure Eight.\footnote{\url{https://www.figure-eight.com/platform/}} We selected a sample of papers from the Semantic Scholar corpus,\footnote{\url{https://semanticscholar.org/}} consisting of papers in general computer science and medicine domains. Citation contexts were extracted using science-parse.\footnote{\url{https://github.com/allenai/science-parse}} The annotators were asked to identify the intent of a citation, and were directed to select among three citation intent options: \method, \result and \background. The annotation interface also included a dummy option \textsc{Other} which helps improve the quality of annotations of other categories. We later removed instances annotated with the \textsc{Other} option from our dataset (less than 1\% of the annotated data), many of which were due to citation contexts which are incomplete or too short for the annotator to infer the citation intent.

We used 50 test questions annotated by a domain expert to ensure crowdsource workers were following directions and disqualify annotators with accuracy less than 75\%. Furthermore, crowdsource workers were required to remain on the annotation page (five annotations) for at least ten seconds before proceeding to the next page. Annotations were dynamically collected. The annotations were aggregated along with a confidence score describing the level of agreement between multiple crowdsource workers. The confidence score is the agreement on a single instance weighted by a trust score (accuracy of the annotator on the initial 50 test questions).

To only collect high quality annotations, instances with confidence score of $\le$0.7 were discarded.
In addition, a subset of the dataset with 100 samples was re-annotated by a trained, expert annotator to check for quality, and the agreement rate with crowdsource workers was \textbf{86\%}. Citation contexts were annotated by 850 crowdsource workers who made a total of 29,926 annotations and individually made between 4 and 240 annotations. Each sentence was annotated, on average, 3.74 times. This resulted in a total 9,159 crowdsourced instances which were divided to training and validation sets with 90\% of the data used for the training set.
In addition to the crowdsourced data, a separate test set of size 1,861 was annotated by a trained, expert annotator to ensure high quality of the dataset.

\subsection{Data for scaffold tasks}

For the first scaffold (citation worthiness), we sample sentences from papers and consider the sentences with citations as positive labels. We also remove the citation markers from those sentences such as numbered citations (e.g., [1]) or name-year combinations (e.g, Lee et al (2012)) to not make the second task artificially easy by only detecting citation markers.
For the second scaffold (citation section title), respective to each test dataset, we sample citations from the ACL-ARC corpus and Semantic Scholar corpus\footnote{\url{https://semanticscholar.org/}} and extract the citation context as well as their corresponding sections. We manually define regular expression patterns mappings to normalized section titles: {``introduction'', ``related work'', ``method'', ``experiments'', ``conclusion''}. Section titles which did not map to any of the aforementioned titles were excluded from the dataset.
Overall, the size of the data for scaffold tasks on the ACL-ARC dataset is about 47K (section title scaffold) and 50K (citation worthiness) while on \ourdata is about 91K and 73K for section title and citation worthiness scaffolds, respectively.

\section{Experiments}
\label{sec:experiments}

\subsection{Implementation}

We implement our proposed scaffold framework using the AllenNLP library \cite{Gardner2017AllenNLP}. For word representations, we use 100-dimensional GloVe vectors \cite{pennington2014glove} trained on a corpus of 6B tokens from Wikipedia and Gigaword. For contextual representations, we use ELMo vectors released by \citet{Peters2018DeepCW}\footnote{\url{https://allennlp.org/elmo}} with output dimension size of 1,024 which have been trained on a dataset of 5.5B tokens. We use a single-layer BiLSTM with a hidden dimension size of 50 for each direction\footnote{Experiments with other types of RNNs such as BiGRUs and more layers showed similar or slightly worst performance}. For each of scaffold tasks, we use a single-layer MLP with 20 hidden nodes , ReLU \cite{nair2010rectified} activation and a Dropout rate \cite{srivastava2014dropout} of 0.2 between the hidden and input layers.
The hyperparameters $\lambda_i$ are tuned for best performance on the validation set of the respective datasets using a 0.0 to 0.3 grid search. For example, the following hyperparameters are used for the ACL-ARC. Citation worthiness saffold: $\lambda_2{=}0.08$, $\lambda_3{=}0$, section title scaffold: $\lambda_3{=}0.09$, $\lambda_2{=}0$; both scaffolds: $\lambda_2{=}0.1$, $\lambda_3{=}0.05$. Batch size is 8 for \acldata dataset and 32 for \ourdata dataset (recall that \ourdata is larger than \acldata). We use Beaker\footnote{Beaker is a collaborative platform for reproducible research (\url{https://github.com/allenai/beaker})} for running the experiments. On the smaller dataset, our best model takes approximately 30 minutes per epoch to train (training time without ELMo is significantly faster).
It is known that multiple runs of probabilistic deep learning models can have variance in overall scores \cite{Reimers2017ReportingSD}\footnote{Some CuDNN methods are non-deterministic and the rest are only deterministic under the same underlying hardware. See \url{https://docs.nvidia.com/deeplearning/sdk/pdf/cuDNN-Developer-Guide.pdf}}. We control this by setting random-number generator seeds; the reported overall results are average of multiple runs with different random seeds.
To facilitate reproducibility, we release our code, data, and trained models.\footnote{\url{https://github.com/allenai/scicite}}

\subsection{Baselines}
\label{subsec:comparison}

We compare our results to several baselines including the model with state-of-the-art performance on the \acldata dataset.
\begin{itemize}[leftmargin=6pt]
\setlength\itemsep{6pt}

    \item \textit{BiLSTM Attention (with and without ELMo)}. This baseline uses a similar architecture to our proposed neural multitask learning framework, except that it only optimizes the network for the main loss regarding the citation intent classification ($\calL_1$) and does not include the structural scaffolds.
    We experiment with two variants of this model: with and without using the contextualized word vector representations (ELMo) of \citet{Peters2018DeepCW}.
    This baseline is useful for evaluating the effect of adding scaffolds in controlled experiments.

    \item \textit{\citet{jurgens2018}}. To make sure our results are competitive with state-of-the-art results on this task, we also compare our model to \citet{jurgens2018} which has the best reported results on the \acldata dataset. \citet{jurgens2018} incorporate a variety of features, ranging from pattern-based features to topic-modeling features, to citation graph features. They also incorporate section titles and relative section position in the paper as features.
    Our implementation of this model achieves a macro-averaged F1 score of 0.526 using 10-fold cross-validation, which is in line with the highest reported results in \citet{jurgens2018}: 0.53 using leave-one-out cross validation. We were not able to use leave-one-out cross validation in our experiments since it is impractical to re-train each variant of our deep learning models thousands of times. Therefore, we opted for a standard setup of stratified train/validation/test data splits with 85\% data used for training and the rest equally split between validation and test.

\end{itemize}

\begin{table}[]
\scriptsize
\centering
\setlength{\tabcolsep}{5pt}
\renewcommand{\arraystretch}{1.2}
\begin{tabular}{@{}llr@{}}
\toprule
 & Model & macro F1 \\ \midrule
\multirow{3}{*}{\rotatebox[origin=c]{90}{\tiny{Baselines}}} & BiLSTM-Attn & 51.8 \\
 & BiLSTM-Attn w/ ELMo & 54.3 \\
 & Previous SOTA \cite{jurgens2018} & 54.6 \\ \midrule
\multirow{3}{*}{\rotatebox[origin=c]{90}{\tiny{This work}}} & BiLSTM-Attn + section title scaffold & 56.9 \\
 & BiLSTM-Attn + citation worthiness scaffold & 56.3 \\
 & BiLSTM-Attn + both scaffolds & 63.1 \\
 & BiLSTM-Attn w/ ELMo + both scaffolds & \bf{67.9} \\
 \bottomrule
\end{tabular}

\caption{\small{Results on the \acldata citations dataset.}}
\label{tab:results-jurgens-data}
\end{table}

\subsection{Results}
\label{subsec:results}

Our main results for the \acldata dataset \cite{jurgens2018} is shown in Table~\ref{tab:results-jurgens-data}. We observe that our scaffold-enhanced models achieve clear improvements over the state-of-the-art approach on this task.
Starting with the `BiLSTM-Attn' baseline with a macro F1 score of 51.8, adding the  first scaffold task in `BiLSTM-Attn + section title scaffold' improves the F1 score to 56.9 ($\Delta{=}5.1$).
Adding the second scaffold in `BiLSTM-Attn + citation worthiness scaffold' also results in similar improvements: 56.3 ($\Delta{=}4.5$).
When both scaffolds are used simultaneously in `BiLSTM-Attn + both scaffolds', the F1 score further improves to 63.1 ($\Delta{=}11.3$), suggesting that the two tasks provide complementary signal that is useful for citation intent prediction.

The best result is achieved when we also add ELMo vectors \cite{Peters2018DeepCW} to the input representations in `BiLSTM-Attn w/ ELMo + both scaffolds', achieving an F1 of 67.9, a major improvement from the previous state-of-the-art results of \citet{jurgens2018} 54.6 ($\Delta{=}13.3$).
We note that the scaffold tasks provide major contributions on top of the ELMo-enabled baseline ($\Delta{=}$13.6), demonstrating the efficacy of using structural scaffolds for citation intent prediction.
We note that these results were obtained without using hand-curated features or additional linguistic resources as used in  \citet{jurgens2018}. We also experimented with adding features used in \citet{jurgens2018} to our best model and not only we did not see any improvements, but we observed at least 1.7\% decline in performance. This suggests that these additional manual features do not provide the model with any additional useful signals beyond what the model already learns from the data.

\setlength{\dashlinedash}{0.5pt}
\setlength{\dashlinegap}{1.0pt}
\setlength{\arrayrulewidth}{0.1pt}

\begin{table}[]
\scriptsize
\centering
\setlength{\tabcolsep}{5pt}
\renewcommand{\arraystretch}{1.2}
\begin{tabular}{@{}llr@{}}
\toprule
 & Model & macro F1 \\ \midrule
\multirow{3}{*}{\rotatebox[origin=c]{90}{\tiny{Baselines}}} & BiLSTM-Attn & 77.2 \\
 & BiLSTM-Attn w/ ELMo & 82.6 \\
 & Previous SOTA \cite{jurgens2018} & 79.6 \\ \midrule
\multirow{3}{*}{\rotatebox[origin=c]{90}{\tiny{This work}}} & BiLSTM-Attn + section title scaffold & 77.8 \\
 & BiLSTM-Attn + citation worthiness scaffold & 78.1 \\
 & BiLSTM-Attn + both scaffolds & 79.1 \\
 & BiLSTM-Attn w/ ELMo + both scaffolds & \bf{84.0} \\
 \bottomrule
\end{tabular}
\caption{\small{Results on the \ourdata dataset.}}
\label{tab:results-our-data}
\end{table}

\begin{table*}[!htbp]
\tiny
\setlength{\tabcolsep}{2.6pt}
\renewcommand{\arraystretch}{1.4}
\centering
\begin{tabular}{@{}lrrrrrrrrrrrrrrrrrrrrr@{}}
\toprule
Category (\# instances) & \multicolumn{3}{c}{Background (71)} & \multicolumn{3}{c}{Compare (25)} & \multicolumn{3}{c}{Extension (5)} & \multicolumn{3}{c}{Future (5)} & \multicolumn{3}{c}{Motivation (7)} & \multicolumn{3}{c}{Use (26)} & \multicolumn{3}{c}{Average (Macro)} \\
\cmidrule(lr){2-4}\cmidrule(lr){5-7}\cmidrule(lr){8-10}\cmidrule(lr){11-13}\cmidrule(lr){14-16}\cmidrule(lr){17-19}\cmidrule(lr){20-22}
 & P & R & F1 & P & R & F1 & P & R & F1 & P & R & F1 & P & R & F1 & P & R & F1 & P & R & F1 \\ \midrule
BiLSTM-Attn & 78.6 & 77.5 & 78.0 & 44.8 & 52.0 & 48.1 & 50.0 & 40.0 & 44.4 & 33.3 & 40.0 & 36.4 & 50.0 & 28.6 & 36.4 & 65.4 & 65.4 & 65.4 & 53.7 & 50.6 & 51.5 \\
BiLSTM-Attn w/ ELMo & 76.5 & 87.3 & 81.6 & 59.1 & 52.0 & 55.3 & 66.7 & 40.0 & 50.0 & 33.3 & 40.0 & 36.4 & 50.0 & 28.6 & 36.4 & 69.6 & 61.5 & 65.3 & 59.2 & 51.6 & 54.2 \\
Previous SOTA \cite{jurgens2018} & 75.6 & 87.3 & 81.1 & 70.6 & 48.0 & 57.1 & 66.7 & 40.0 & 50.0 & 50.0 & 20.0 & 28.6 & 75.0 & \bf{42.9} & \bf{54.6} & 51.6 & 61.5 & 56.1 & 64.9 & 49.9 & 54.6 \\ \hdashline
BiLSTM-Attn+section title scaffold & 77.2 & 85.9 & 81.3 & 53.8 & 56.0 & 54.9 & \bf{100.0} & 40.0 & 57.1 & 33.3 & 40.0 & 36.4 & 50.0 & 28.6 & 36.4 & \bf{81.8} & \bf{69.2} & \bf{75.0} & 66.0 & 53.3 & 56.9 \\
BiLSTM-Attn+citation worthiness scaffold & 77.1 & 90.1 & 83.1 & 59.1 & 52.0 & 55.3 & \bf{100.0} & 40.0 & 57.1 & 28.6 & 40.0 & 33.3 & 50.0 & 28.6 & 36.4 & 81.0 & 65.4 & 72.3 & 66.0 & 52.7 & 56.3 \\
BiLSTM-Attn+both scaffolds & \bf{77.6} & \bf{93.0} & \bf{84.6} & 65.0 & 52.0 & 57.8 & \bf{100.0} & \bf{60.0} & \bf{75.0} & 40.0 & 40.0 & 40.0 & 75.0 & \bf{42.9} & 54.5 & 72.7 & 61.5 & 66.7 & 71.7 & 58.2 & 63.1 \\
BiLSTM-Attn+both scaffolds /w ELMo & 75.9 & \bf{93.0} & 83.5 & \bf{80.0} & \bf{64.0} & \bf{71.1} & 75.0 & \bf{60.0} & 66.7 & \bf{75.0} & \bf{60.0} & \bf{66.7} & \bf{100.0} & 28.6 & 44.4 & \bf{81.8} & \bf{69.2} & \bf{75.0} & \bf{81.3} & \bf{62.5} & \bf{67.9} \\ \bottomrule
\end{tabular}
\caption{\small{Detailed per category classification results on \acldata dataset.}}
\label{tab:results-per-category}
\end{table*}

\begin{table*}[!htbp]
\tiny
\setlength{\tabcolsep}{6.3pt}
\renewcommand{\arraystretch}{1.4}
\centering
\begin{tabular}{@{}llllllllllllll@{}}
\toprule
Category (\# instances) & \multicolumn{3}{c}{Background (1,014)} & \multicolumn{3}{c}{Method (613)} & \multicolumn{3}{c}{Result (260)} & \multicolumn{3}{c}{Average (Macro)} \\
\cmidrule(lr){2-4}\cmidrule(lr){5-7}\cmidrule(lr){8-10}\cmidrule(lr){11-13}
 & P & R & F1 & P & R & F1 & P & R & F1 & P & R & F1 \\ \midrule
BiLSTM-Attn & 82.2 & 83.2 & 82.7 & 80.7 & 74.4 & 77.4 & 67.1 & 76.2 & 71.4 & 76.7 & 77.9 & 77.2  \\
BiLSTM-Attn w/ ELMo & \bf{86.6} & 87 & 86.8 & 87.2 & 79.1 & 83.0 & 71.5 & \bf{85.8} & 78.0 & 81.8 & \bf{84.0} & 82.6  \\
Previous SOTA \cite{jurgens2018} & 77.9 & \bf{92.9} & 84.7 & \bf{91.5} & 63.1 & 74.7 & 79.1 & 77.3 & 78.2 & 82.8 & 77.8 & 79.2 \\
\hdashline
BiLSTM-Attn + section title scaffold & 81.3 & 86.0 & 83.6 & 85.3 & 68.8 & 76.2 & 66.8 & 81.9 & 73.6 & 77.8 & 78.9 & 77.8 \\
BiLSTM-Attn + citation worthiness scaffold & 82.9 & 84.8 & 83.8 & 84.6 & 73.2 & 78.5 & 65.4 & 80.0 & 72.0 & 77.6 & 79.3 & 78.1 \\
BiLSTM-Attn + both scaffolds & 85.4 & 80.8 & 83.0 & 78.6 & 80.4 & 79.5 & 69.8 & 80.8 & 74.9 & 77.9 & 80.7 & 79.1 \\
BiLSTM-Attn w/ ELMo + both scaffolds & 85.4 & 90.3 & \bf{87.8} & 89.5 & \bf{80.8} & \bf{84.9} & \bf{79.3} & 79.6 & \bf{79.5} & \bf{84.7} & 83.6 & \bf{84.0} \\ \bottomrule
\end{tabular}
\caption{Detailed per category classification results on the \ourdata dataset. }
\label{tab:results-per-category-scienecite}
\end{table*}

Table~\ref{tab:results-our-data} shows the main results on \ourdata dataset, where we see similar patterns. Each scaffold task improves model performance. Adding both scaffolds results in further improvements. And the best results are obtained by using ELMo representation in addition to both scaffolds.
Note that this dataset is more than five times larger in size than the \acldata, therefore the performance numbers are generally higher and the F1 gains are generally smaller since it is easier for the models to learn optimal parameters utilizing the larger annotated data.
On this dataset, the best baseline is the neural baseline with addition of ELMo contextual vectors achieving an F1 score of 82.6 followed by \citet{jurgens2018}, which is expected because neural models generally achieve higher gains when more training data is available and because \citet{jurgens2018} was not designed with the \ourdata dataset in mind.

The breakdown of results by intent on \acldata and \ourdata datasets is respectively shown in Tables~\ref{tab:results-per-category}~and~\ref{tab:results-per-category-scienecite}. Generally we observe that results on categories with more number of instances are higher. For example on \acldata, the results on the \background category are the highest as this category is the most common. Conversely, the results on the \future category are the lowest. This category has the fewest data points (see distribution of the categories in Table~\ref{tab:data}) and thus it is harder for the model to learn the optimal parameters for correct classification in this category.

\subsection{Analysis}
\label{subsec:analysis-params}

\begin{figure}[]
\centering
   \begin{subfigure}[b]{\linewidth}
   \includegraphics[width=\linewidth]{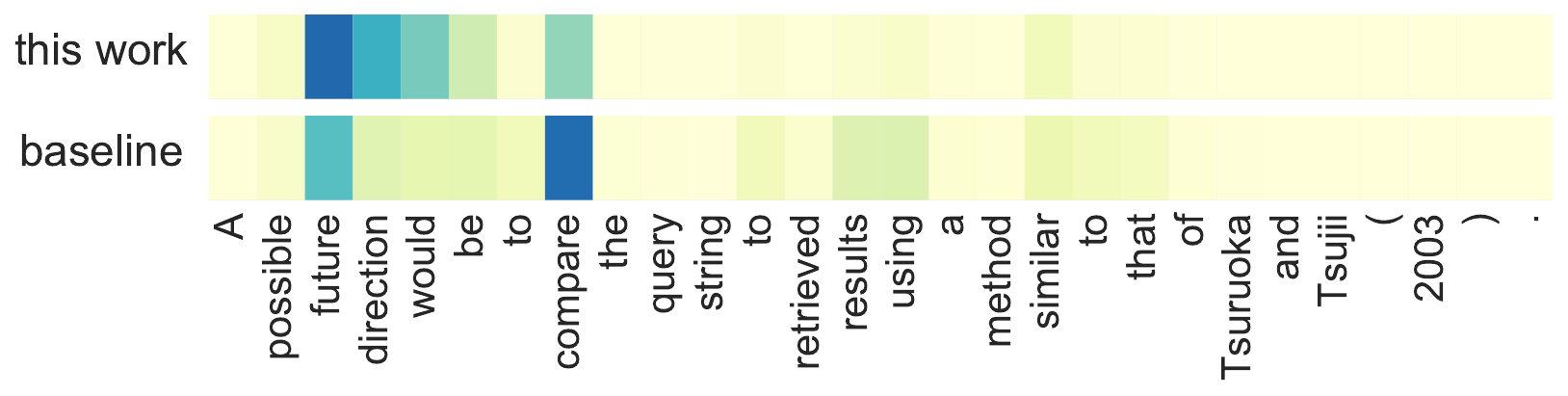}
    \caption{\scriptsize{Example from ACL-ARC: Correct label is \future. Our model correctly predicts it while baseline predicts \compare.}}
   \label{fig:Ng1}
    \end{subfigure}
\\
    \begin{subfigure}[b]{\linewidth}
   \includegraphics[width=\linewidth]{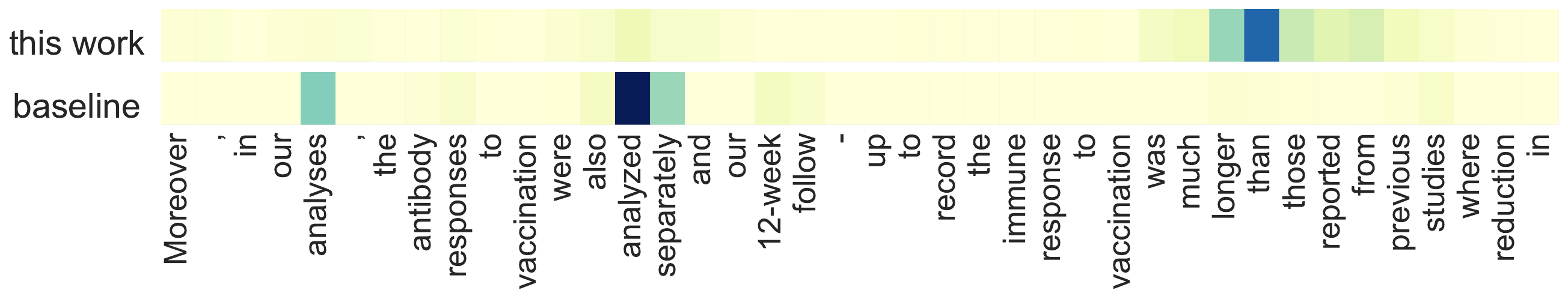}
   \caption{\scriptsize{Example from \ourdata: Correct label is \result; our model correctly predicts it, while baseline considers it as \background.}}
   \label{fig:Ng2}
    \end{subfigure}
\caption{
\small{Visualization of attention weights corresponding to our best scaffold model  compared with the best baseline neural baseline model without scaffolds}.
}
\label{fig:examples}
\end{figure}

To gain more insight into why the scaffolds are helping the model in improved citation intent classification, we examine the attention weights assigned to inputs for our best proposed model (`BiLSTM-Attn w/ ELMo + both scaffolds') compared with the best neural baseline (`BiLSTM-Attn w/ ELMO'). We conduct this analysis for examples from both datasets. Figure~\ref{fig:examples} shows an example input citation along with the horizontal line and the heatmap of attention weights for this input resulting from our model versus the baseline. For first example (\ref{fig:Ng1}) the true label is \future. We observe that our model puts more weight on words surrounding the word ``future'' which is plausible given the true label. On the other hand, the baseline model attends most to the words ``compare'' and consequently incorrectly predicts a \compare label. In second example (\ref{fig:Ng2}) the true label is \result. The baseline incorrectly classifies it as a \background, likely due to attending to another part of the sentence (``analyzed seprately''). Our model correctly classifies this instance by putting more attention weights on words that relate to comparison of the results. This suggests that the our model is more successful in learning optimal parameters for representing the citation text and classifying its respective intent compared with the baseline. Note that the only difference between our model and the neural baseline is inclusion of the structural scaffolds. Therefore, suggesting the effectiveness the scaffolds in informing the main task of relevant signals for citation intent classification.

\setlength{\dashlinedash}{0.5pt}
\setlength{\dashlinegap}{1.0pt}
\setlength{\arrayrulewidth}{0.1pt}

\begin{table}[]
\tiny
\centering
\setlength{\tabcolsep}{2pt}
\renewcommand{\arraystretch}{1.9}
\begin{tabular}{@{}p{4.8cm}rr@{}}
\toprule
Example & True & Prediction \\ \midrule
Our work is inspired by the latent left-linking model in (CITATION) and the ILP formulation from (CITATION). & \motivation & \use \\ \hdashline
ASARES is presented in detail in (CITATION) . & \use & \background \\ \hdashline
The advantage of tuning similarity to the application of interest has been shown previously by (CITATION). & \textsc{Compare} & \background \\ \hdashline
One possible direction is to consider linguistically motivated approaches , such as the extraction of syntactic phrase tables as proposed by (CITATION). & \future & \background \\ \hdashline
After the extraction, pruning techniques (CITATION) can be applied to increase the precision of the extraction. & \background & \use \\ \bottomrule
\end{tabular}
\caption{A sample of model's classification errors on \acldata dataset}
\label{tab:example-errors}
\end{table}

\begin{figure}[]
\centering
  \begin{subfigure}[b]{0.49\linewidth}
  \includegraphics[width=0.9\linewidth]{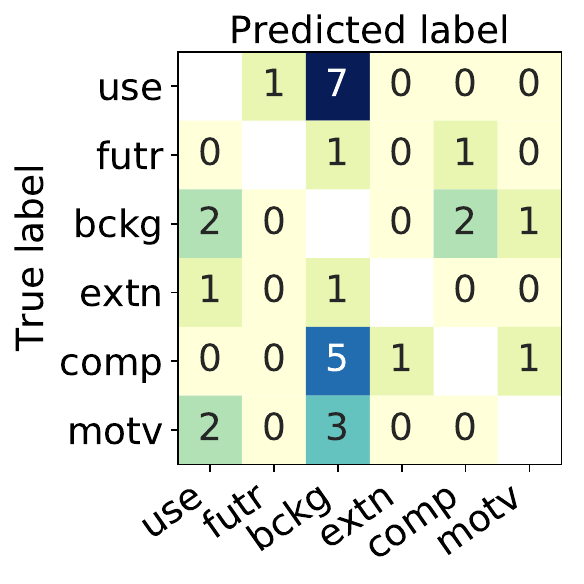}
  \caption{\scriptsize{ACL-ARC (test size: 139)} \label{fig:confusion-subfig1}}
    \end{subfigure}
    \begin{subfigure}[b]{0.49\linewidth}
  \includegraphics[width=0.9\linewidth]{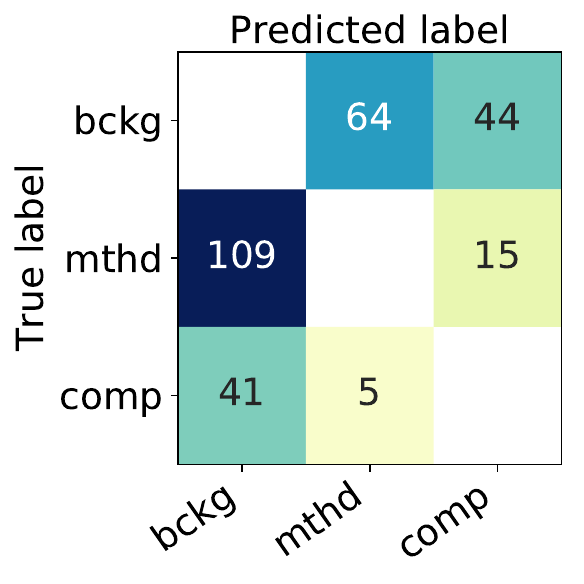}
  \caption{\scriptsize{\ourdata (test size: 1,861)} \label{fig:confusion-subfig2} }
    \end{subfigure}
\caption{
Confusion matrix showing classification errors of our best model on two datasets. The diagonal is masked to bring focus only on errors.
}
\label{fig:confusion}
\vspace{-10pt}
\end{figure}

\paragraph{Error analysis.}
We next investigate errors made by our best model (Figure~\ref{fig:confusion} plots classification errors). One general error pattern is that the model has more tendency to make false positive errors in the \background category likely due to this category dominating both datasets. It's interesting that for the \acldata dataset some prediction errors are due to the model failing to properly differentiate the \use category with \background. We found out that some of these errors would have been possibly prevented by using additional context. Table~\ref{tab:example-errors} shows a sample of such classification errors. For the citation in the first row of the table, the model is likely distracted by ``model in (citation)'' and ``ILP formulation from (citation)'' deeming the sentence is referring to the use of another method from a cited paper and it misses the first part of the sentence describing the motivation. This is likely due to the small number of training instances in the \motivation category, preventing the model to learn such nuances. For the examples in the second and third row, it is not clear if it is possible to make the correct prediction without additional context. And similarly in the last row the instance seems ambiguous without accessing to additional context. Similarly as shown in Figure~\ref{fig:confusion-subfig1} two of \future labels are wrongly classified. One of them is illustrated in the forth row of Table~\ref{tab:example-errors} where perhaps additional context could have helped the model in identifying the correct label. One possible way to prevent this type of errors, is to provide the model with an additional input, modeling the extended surrounding context. We experimented with encoding the extended surrounding context using a BiLSTM and concatenating it with the main citation context vector (z), but it resulted in a large decline in overall performance likely due to the overall noise introduced by the additional context. A possible future work is to investigate alternative effective approaches for incorporating the surrounding extended context.

\section{Related Work}
\label{sec:related}

There is a large body of work studying the intent of citations and devising categorization systems \cite{stevens1965statistical,moravcsik1975some,garzone2000towards,white2004citation,ahmed2004highly,teufel2006function,agarwal2010automatically,dong2011ensemble}. Most of these efforts provide citation categories that are too fine-grained, some of which rarely occur in papers. Therefore, they are hardly useful for automated analysis of scientific publications. To address these problems and to unify previous efforts, in a recent work, \citet{jurgens2018} proposed a six category system for citation intents. In this work, we focus on two schemes: (1) the scheme proposed by \citet{jurgens2018} and (2) an additional, more coarse-grained general-purpose category system that we propose (details in \S\ref{sec:data}). Unlike other schemes that are domain-specific, our scheme is general and naturally fits in scientific discourse in multiple domains.

Early works in automated citation intent classification were based on rule-based systems (e.g., \cite{garzone2000towards,pham2003new}). Later, machine learning methods based on linguistic patterns and other hand-engineered features from citation context were found to be effective. For example, \citet{teufel2006function} proposed use of ``cue phrases'', a set of expressions
that talk about the act of presenting research
in a paper. \citet{abu2013purpose} relied on lexical, structural, and syntactic features and a linear SVM for classification. Researchers have also investigated methods of finding cited spans in the cited papers. Examples include feature-based methods \cite{Cohan2015MatchingCT}, domain-specific knowledge \cite{Cohan2017ContextualizingCF}, and a recent CNN-based model for joint prediction of cited spans and citation function \cite{Su2018NeuralML}. We also experimented with CNNs but found the attention BiLSTM model to work significantly better.
\citet{jurgens2018} expanded all pre-existing feature-based efforts on citation intent classification by proposing a comprehensive set of engineered features, including boostrapped patterns, topic modeling, dependency-based, and metadata features for the task. We argue that we can capture necessary information from the citation context using a data driven method, without the need for hand-engineered domain-dependent features or external resources. We propose a novel scaffold neural model for citation intent classification to incorporate structural information of scientific discourse into citations, borrowing the ``scaffold'' terminology from \citet{Swayamdipta2018SyntacticSF} who use auxiliary syntactic tasks for semantic problems.

\section{Conclusions and future work}
\label{sec:conclusion}

In this work, we show that structural properties related to scientific discourse can be effectively used to inform citation intent classification. We propose a multitask learning framework with two auxiliary tasks (predicting section titles and citation worthiness) as two scaffolds related to the main task of citation intent prediction. Our model achieves state-of-the-art result (F1 score of 67.9\%) on the \acldata dataset with 13.3 absolute increase over the best previous results. We additionally introduce \ourdata, a new large dataset of citation intents and also show the effectiveness of our model on this dataset. Our dataset, unlike existing datasets that are designed based on a specific domain, is more general and fits in scientific discourse from multiple scientific domains.

We demonstrate that carefully chosen auxiliary tasks that are inherently relevant to a main task can be leveraged to improve the performance on the main task. An interesting line of future work is to explore the design of such tasks or explore the properties or similarities between the auxiliary and the main tasks. Another relevant line of work is adapting our model to other domains containing documents with similar linked structured such as Wikipedia articles.
Future work may benefit from replacing ELMo with other types of contextualized representations such as BERT in our scaffold model. For example, at the time of finalizing the camera ready version of this paper, \citet{Beltagy2019SciBERTPC} showed that a BERT contextualized representation model \cite{Devlin2018BERTPO} trained on scientific text can achieve promising results on the \ourdata dataset.

\section*{Acknowledgments}

We thank Kyle Lo, Dan Weld, and Iz Beltagy for helpful discussions, Oren Etzioni for feedback on the paper, David Jurgens for helping us with their ACL-ARC dataset and reproducing their results, and the three anonymous reviewers for their comments and suggestions. Computations on \url{beaker.org} were supported in part by credits from Google Cloud.

\bibliography{naaclhlt2019}

\begin{thebibliography}{33}
\expandafter\ifx\csname natexlab\endcsname\relax\def\natexlab#1{#1}\fi

\bibitem[{Abu-Jbara et~al.(2013)Abu-Jbara, Ezra, and Radev}]{abu2013purpose}
Amjad Abu-Jbara, Jefferson Ezra, and Dragomir Radev. 2013.
\newblock Purpose and polarity of citation: Towards nlp-based bibliometrics.
\newblock In \emph{NAACL-HLT}.

\bibitem[{Agarwal et~al.(2010)Agarwal, Choubey, and
  Yu}]{agarwal2010automatically}
Shashank Agarwal, Lisha Choubey, and Hong Yu. 2010.
\newblock Automatically classifying the role of citations in biomedical
  articles.
\newblock In \emph{AMIA Annual Symposium Proceedings}, volume 2010, page~11.
  American Medical Informatics Association.

\bibitem[{Ahmed et~al.(2004)Ahmed, Johnson, Oppenheim, and
  Peck}]{ahmed2004highly}
Tanzila Ahmed, Ben Johnson, Charles Oppenheim, and Catherine Peck. 2004.
\newblock Highly cited old papers and the reasons why they continue to be
  cited. part ii., the 1953 watson and crick article on the structure of dna.
\newblock \emph{Scientometrics}, 61(2):147--156.

\bibitem[{Beltagy et~al.(2019)Beltagy, Cohan, and Lo}]{Beltagy2019SciBERTPC}
Iz~Beltagy, Arman Cohan, and Kyle Lo. 2019.
\newblock Scibert: Pretrained contextualized embeddings for scientific text.
\newblock \emph{CoRR}, abs/1903.10676.

\bibitem[{Bird et~al.(2008)Bird, Dale, Dorr, Gibson, Joseph, Kan, Lee, Powley,
  Radev, and Tan}]{Bird2008TheAA}
Steven Bird, Robert Dale, Bonnie~J. Dorr, Bryan~R. Gibson, Mark~Thomas Joseph,
  Min-Yen Kan, Dongwon Lee, Brett Powley, Dragomir~R. Radev, and Yee~Fan Tan.
  2008.
\newblock The acl anthology reference corpus: A reference dataset for
  bibliographic research in computational linguistics.
\newblock In \emph{LREC}.

\bibitem[{Caruana(1997)}]{Caruana1997MultitaskL}
Rich Caruana. 1997.
\newblock Multitask learning.
\newblock \emph{Machine Learning}, 28:41--75.

\bibitem[{Cohan and Goharian(2015)}]{cohan2015summarization}
Arman Cohan and Nazli Goharian. 2015.
\newblock Scientific article summarization using citation-context and
  article{'}s discourse structure.
\newblock In \emph{EMNLP}.

\bibitem[{Cohan and Goharian(2017)}]{Cohan2017ContextualizingCF}
Arman Cohan and Nazli Goharian. 2017.
\newblock Contextualizing citations for scientific summarization using word
  embeddings and domain knowledge.
\newblock In \emph{SIGIR}.

\bibitem[{Cohan et~al.(2015)Cohan, Soldaini, and
  Goharian}]{Cohan2015MatchingCT}
Arman Cohan, Luca Soldaini, and Nazli Goharian. 2015.
\newblock Matching citation text and cited spans in biomedical literature: a
  search-oriented approach.
\newblock In \emph{HLT-NAACL}.

\bibitem[{Devlin et~al.(2018)Devlin, Chang, Lee, and
  Toutanova}]{Devlin2018BERTPO}
Jacob Devlin, Ming-Wei Chang, Kenton Lee, and Kristina Toutanova. 2018.
\newblock Bert: Pre-training of deep bidirectional transformers for language
  understanding.
\newblock \emph{CoRR}, abs/1810.04805.

\bibitem[{Dong and Sch{\"a}fer(2011)}]{dong2011ensemble}
Cailing Dong and Ulrich Sch{\"a}fer. 2011.
\newblock Ensemble-style self-training on citation classification.
\newblock In \emph{IJCNLP}.

\bibitem[{Gardner et~al.(2018)Gardner, Grus, Neumann, Tafjord, Dasigi, Liu,
  Peters, Schmitz, and Zettlemoyer}]{Gardner2017AllenNLP}
Matt Gardner, Joel Grus, Mark Neumann, Oyvind Tafjord, Pradeep Dasigi,
  Nelson~F. Liu, Matthew~E. Peters, Michael Schmitz, and Luke~S. Zettlemoyer.
  2018.
\newblock Allennlp: A deep semantic natural language processing platform.
\newblock \emph{CoRR}, abs/1803.07640.

\bibitem[{Garzone and Mercer(2000)}]{garzone2000towards}
Mark Garzone and Robert~E Mercer. 2000.
\newblock Towards an automated citation classifier.
\newblock In \emph{Conference of the Canadian Society for Computational Studies
  of Intelligence}, pages 337--346. Springer.

\bibitem[{Hochreiter and Schmidhuber(1997)}]{Hochreiter1997LongSM}
Sepp Hochreiter and J{\"u}rgen Schmidhuber. 1997.
\newblock Long short-term memory.
\newblock \emph{Neural Computation}.

\bibitem[{Jurgens et~al.(2018)Jurgens, Kumar, Hoover, McFarland, and
  Jurafsky}]{jurgens2018}
David Jurgens, Srijan Kumar, Raine Hoover, Dan McFarland, and Dan Jurafsky.
  2018.
\newblock \href {https://transacl.org/ojs/index.php/tacl/article/view/1266}
  {Measuring the evolution of a scientific field through citation frames}.
\newblock \emph{TACL}, 6:391--406.

\bibitem[{Leydesdorff(1998)}]{leydesdorff1998theories}
Loet Leydesdorff. 1998.
\newblock Theories of citation?
\newblock \emph{Scientometrics}.

\bibitem[{Li and Ho(2008)}]{Li2008}
Zhi Li and Yuh-Shan Ho. 2008.
\newblock \href {https://doi.org/10.1007/s11192-007-1838-1} {Use of citation
  per publication as an indicator to evaluate contingent valuation research}.
\newblock \emph{Scientometrics}.

\bibitem[{Luukkonen(1992)}]{luukkonen1992scientists}
Terttu Luukkonen. 1992.
\newblock Is scientists' publishing behaviour rewardseeking?
\newblock \emph{Scientometrics}.

\bibitem[{Moravcsik and Murugesan(1975)}]{moravcsik1975some}
Michael~J Moravcsik and Poovanalingam Murugesan. 1975.
\newblock Some results on the function and quality of citations.
\newblock \emph{Social studies of science}, 5(1):86--92.

\bibitem[{Nair and Hinton(2010)}]{nair2010rectified}
Vinod Nair and Geoffrey~E Hinton. 2010.
\newblock Rectified linear units improve restricted boltzmann machines.
\newblock In \emph{Proceedings of the 27th international conference on machine
  learning (ICML-10)}, pages 807--814.

\bibitem[{Pennington et~al.(2014)Pennington, Socher, and
  Manning}]{pennington2014glove}
Jeffrey Pennington, Richard Socher, and Christopher Manning. 2014.
\newblock Glove: Global vectors for word representation.
\newblock In \emph{EMNLP}, pages 1532--1543.

\bibitem[{Peters et~al.(2018)Peters, Neumann, Iyyer, Gardner, Clark, Lee, and
  Zettlemoyer}]{Peters2018DeepCW}
Matthew~E. Peters, Mark Neumann, Mohit Iyyer, Matt Gardner, Christopher Clark,
  Kenton Lee, and Luke~S. Zettlemoyer. 2018.
\newblock Deep contextualized word representations.
\newblock In \emph{NAACL-HLT}.

\bibitem[{Pham and Hoffmann(2003)}]{pham2003new}
Son~Bao Pham and Achim Hoffmann. 2003.
\newblock A new approach for scientific citation classification using cue
  phrases.
\newblock In \emph{Australasian Joint Conference on Artificial Intelligence},
  pages 759--771. Springer.

\bibitem[{Reimers and Gurevych(2017)}]{Reimers2017ReportingSD}
Nils Reimers and Iryna Gurevych. 2017.
\newblock Reporting score distributions makes a difference: Performance study
  of lstm-networks for sequence tagging.
\newblock In \emph{EMNLP}.

\bibitem[{Ritchie(2009)}]{ritchie2009citation}
Anna Ritchie. 2009.
\newblock Citation context analysis for information retrieval.
\newblock Technical report, University of Cambridge, Computer Laboratory.

\bibitem[{Small(2018)}]{small2018}
Henry Small. 2018.
\newblock \href {https://doi.org/https://doi.org/10.1016/j.joi.2018.03.007}
  {Characterizing highly cited method and non-method papers using citation
  contexts: The role of uncertainty}.
\newblock \emph{Journal of Informetrics}, 12(2):461 -- 480.

\bibitem[{Srivastava et~al.(2014)Srivastava, Hinton, Krizhevsky, Sutskever, and
  Salakhutdinov}]{srivastava2014dropout}
Nitish Srivastava, Geoffrey Hinton, Alex Krizhevsky, Ilya Sutskever, and Ruslan
  Salakhutdinov. 2014.
\newblock Dropout: a simple way to prevent neural networks from overfitting.
\newblock \emph{The Journal of Machine Learning Research}, 15(1):1929--1958.

\bibitem[{Stevens and Giuliano(1965)}]{stevens1965statistical}
Mary~Elizabeth Stevens and Vincent~Edward Giuliano. 1965.
\newblock \emph{Statistical Association Methods for Mechanized Documentation:
  Symposium Proceedings, Washington, 1964}, volume 269.
\newblock US Government Printing Office.

\bibitem[{Su et~al.(2018)Su, Prasad, Kan, and Sugiyama}]{Su2018NeuralML}
Xuan Su, Animesh Prasad, Min-Yen Kan, and Kazunari Sugiyama. 2018.
\newblock Neural multi-task learning for citation function and provenance.
\newblock \emph{CoRR}, abs/1811.07351.

\bibitem[{Swayamdipta et~al.(2018)Swayamdipta, Thomson, Lee, Zettlemoyer, Dyer,
  and Smith}]{Swayamdipta2018SyntacticSF}
Swabha Swayamdipta, Sam Thomson, Kenton Lee, Luke~S. Zettlemoyer, Chris Dyer,
  and Noah~A. Smith. 2018.
\newblock Syntactic scaffolds for semantic structures.
\newblock In \emph{EMNLP}.

\bibitem[{Teufel et~al.(2006)Teufel, Siddharthan, and
  Tidhar}]{teufel2006function}
Simone Teufel, Advaith Siddharthan, and Dan Tidhar. 2006.
\newblock \href {http://dl.acm.org/citation.cfm?id=1610075.1610091} {Automatic
  classification of citation function}.
\newblock In \emph{EMNLP}, EMNLP '06, pages 103--110, Stroudsburg, PA, USA.
  Association for Computational Linguistics.

\bibitem[{White(2004)}]{white2004citation}
Howard~D White. 2004.
\newblock Citation analysis and discourse analysis revisited.
\newblock \emph{Applied linguistics}, 25(1):89--116.

\bibitem[{Zeiler(2012)}]{zeiler2012adadelta}
Matthew~D Zeiler. 2012.
\newblock Adadelta: an adaptive learning rate method.
\newblock \emph{arXiv preprint arXiv:1212.5701}.

\end{thebibliography}
\bibliographystyle{acl_natbib}

\end{document}